\documentclass[runningheads]{llncs}

\usepackage{array}             
\usepackage{mwe}               
\usepackage{multirow}          
\usepackage{float}             
\usepackage{booktabs}          
\usepackage{colortbl}          
\usepackage{ulem}              
\usepackage{hyperref}          
\usepackage{url}               
\usepackage{arydshln}          
\usepackage{hhline}            
\usepackage{wrapfig}           
\usepackage{lipsum}            
\usepackage{booktabs}          
\usepackage{xcolor}            
\usepackage{caption}           
\usepackage[T1]{fontenc}       
\usepackage{graphicx}          
\usepackage{verbatim}          
\usepackage{subcaption}        

\newcommand{\eva}{\texttt{eva}\xspace}
\newcommand{\coralbay}{\texttt{CoralBay}\xspace}

\begin{document}

\title{\coralbay: A Self-Supervised CT Foundation Model}

\author{
   Ioannis Gatopoulos \and
   Nicolas Känzig \and
   Sebastian Otálora \and
   Fei Tang
}


\institute{
    kaiko.ai\\
    \email{\{ioannis,nicolas,sebastian,fei\}@kaiko.ai}
}

\maketitle 

\begin{abstract}
Self-supervised learning has enabled large-scale pre-training on 2D natural images, producing general-purpose visual representations that transfer effectively across tasks. However, many medical imaging modalities, such as CT scans, are inherently three-dimensional and differ fundamentally from natural images in both structure and semantics. Volumetric modalities capture spatial continuity, organ anatomy, and intensity-based tissue properties (e.g., Hounsfield Units), which are not adequately modeled by 2D pre-training. To bridge this gap, we introduce \coralbay, a self-distillation framework that extends DINO by using a hierarchical 3D Swin backbone and applying self-distillation to concatenated multi-scale features, enabling data-efficient self-supervised learning of rich spatial representations that encode both global semantics and fine-grained local structure. As a result, \coralbay transfers effectively to a wide range of downstream radiological tasks, demonstrating strong and consistent performance across diverse anatomical targets. In addition, we contribute to the open-source \eva framework by introducing a public, reproducible 3D radiology leaderboard that unifies multiple datasets and establishes a standardized benchmark for evaluating volumetric representation learning methods.

\end{abstract}

\section{Introduction}

\begin{figure}[t]
    \centering
    \includegraphics[width=1.\linewidth]{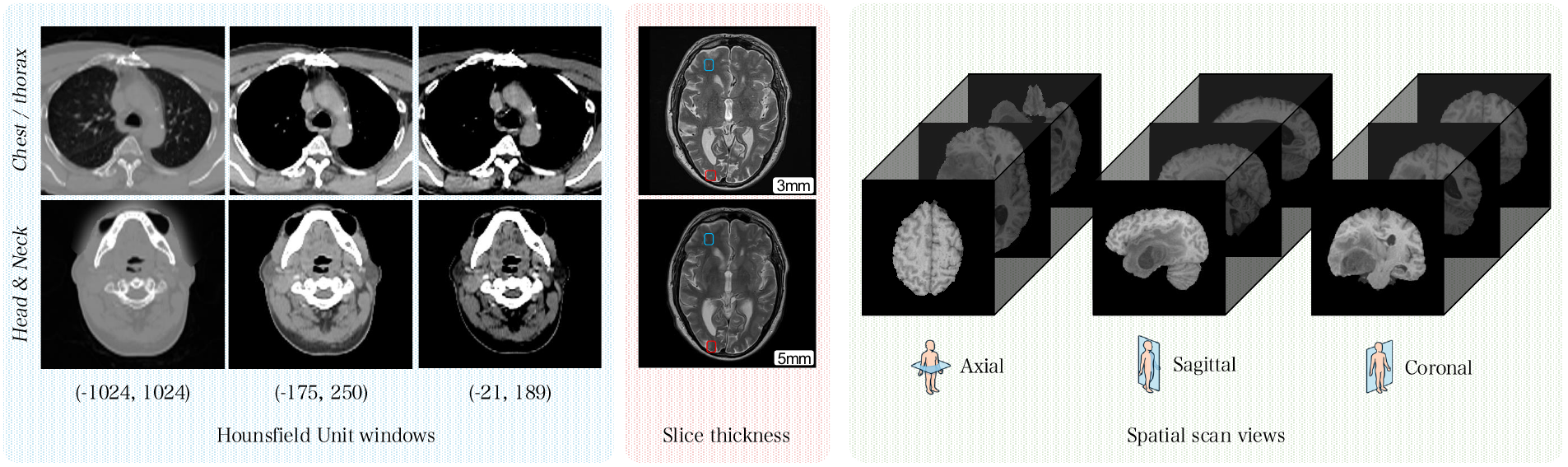}
    \captionsetup{font=small}
    \caption{\textbf{Challenges in 3D CT Data Representation.} \textbf{Left:} The impact of HU windowing on anatomical visibility; narrow windows enhance soft tissue but clip high-density information. \textbf{Center:} Variability in slice thickness leads to the partial volume effect, where thinner slices provide higher spatial resolution while thicker slices increase the signal-to-noise ratio at the cost of blurring. \textbf{Right:} The three standard orthogonal planes which require the model to learn spatially consistent 3D representations across high-dimensional volumes.}
    \label{fig:ct_challenges}
\end{figure}

Despite the success~\cite{he2020momentum,bommasani2021opportunities,radford2021learning} of vision foundation models (FMs), their adoption in 3D medical imaging remains limited due to fundamental differences between natural images and volumetric radiological data, as illustrated in Fig.~\ref{fig:ct_challenges}. Unlike 2D RGB images, CT and MRI modalities consist of high-resolution three-dimensional volumes that encode physical tissue properties. This introduces several domain-specific hurdles:
\begin{itemize}
    \item \textbf{Intensity Variability and Windowing:} Voxel intensities are measured in Hounsfield Units (HU), representing physical density rather than color. As shown in Fig.~\ref{fig:ct_challenges} (left), different HU windows are required to visualize specific tissues; a window optimized for the lungs will completely obscure soft-tissue details in the neck, creating a challenge for models to learn consistent features across varied visualization protocols.
    \item \textbf{Anisotropy and Slice Thickness:} Medical volumes are often non-isometric. Variations in slice thickness (e.g., 3mm vs. 5mm in Fig.~\ref{fig:ct_challenges}, center) exacerbate partial volume effects; thicker slices blur anatomical boundaries and reduce spatial resolution, despite increasing signal-to-noise ratio, potentially introducing systematic bias rather than random noise into the learning process.
    \item \textbf{Spatial Complexity:} The transition from 2D frames to 3D spatial scan views (Axial, Sagittal, Coronal in Fig.~\ref{fig:ct_challenges}, right) requires architectures capable of modeling long-range dependencies across planes while managing a massive memory footprint and severe class imbalances, such as tiny nodules within a vast anatomical volume.
\end{itemize}

To bridge this gap, we propose \coralbay, a self-supervised framework that natively operates on 3D medical volumes. Building on the self-distillation principles of DINO \cite{caron2021dino}, our approach extends them to volumetric data through hierarchical Swin Transformers \cite{liu2021swin}, multi-resolution feature learning, and 3D-specific augmentation strategies. Crucially, our training pipeline is tailored to the physical properties of CT imaging, exposing the model to diverse HU windows and realistic intensity shifts to ensure robustness against acquisition variability. To support reproducibility and benchmarking, we further extend the \eva open-source framework~\cite{gatopoulos2024eva} with a public 3D radiology leaderboard.\vspace{-4pt}

\section{Related Work}

Self-supervised and weakly supervised learning have emerged as central paradigms for visual representation learning, enabling models to leverage large-scale unlabeled data through carefully designed pretext objectives \cite{jing2020self}. This approach is especially valuable in medical imaging, where expert annotations are costly, time-consuming, and constrained by privacy regulations, while vast amounts of unlabeled data are routinely generated in clinical workflows \cite{shen2017deep}.

In natural images, effective self-supervised representation learning relies on objectives that derive supervision directly from the data itself. Cross-modal contrastive methods align visual features with language semantics using paired image-text data, yielding highly transferable embeddings, like Universal Model \cite{Liu_2023}and SuPreM \cite{li2025supervised3dmodelstransfer}, but are constrained by the scale of paired data. In contrast, vision-only approaches learn from intrinsic image structure, either by contrasting augmented views of the same instance \cite{he2020momentum,chen2020simple} or through prototype-level objectives based on clustering and teacher-student self-distillation without explicit negatives \cite{caron2021dino,grill2020bootstrap,oquab2024dinov2learningrobustvisual}. This work aligns with the latter category. Extending these ideas to 3D medical imaging presents additional challenges, including volumetric structure, anisotropic resolution, and modality-specific intensity distributions. Early efforts such as Models Genesis explored context restoration for 3D volumes \cite{zhou2021modelsgenesis}, while later methods introduced tailored 3D pretext tasks like view-based or region-based objectives \cite{taleb20213d,zhang2022self,tang2022self}. VoCo \cite{voco} proposed a volume contrastive learning framework with geometric context priors by contrasting augmented views of the same 3D volume. Despite strong performance, it relies on heuristic view design and may be limited in capturing high-level semantics. Approaches using Swin Transformers as backbones are particularly relevant: MoBY combined MoCo v2 and BYOL on Swin Transformers, and achieved promising results on 2D natural images, especially on dense prediction tasks \cite{tang2022self}, while Swin UNETR introduced hierarchical Swin-based pre-training on proxy tasks, leading to state-of-the-art performance on CT and MRI segmentation \cite{tang2022self,hatamizadeh2022swinunetr}.

\section{Methodology}

\begin{figure}[t]
    \centering
    \scalebox{1.}{
    \begin{subfigure}{1.\textwidth}
    \centering
    \includegraphics[width=1.\linewidth]{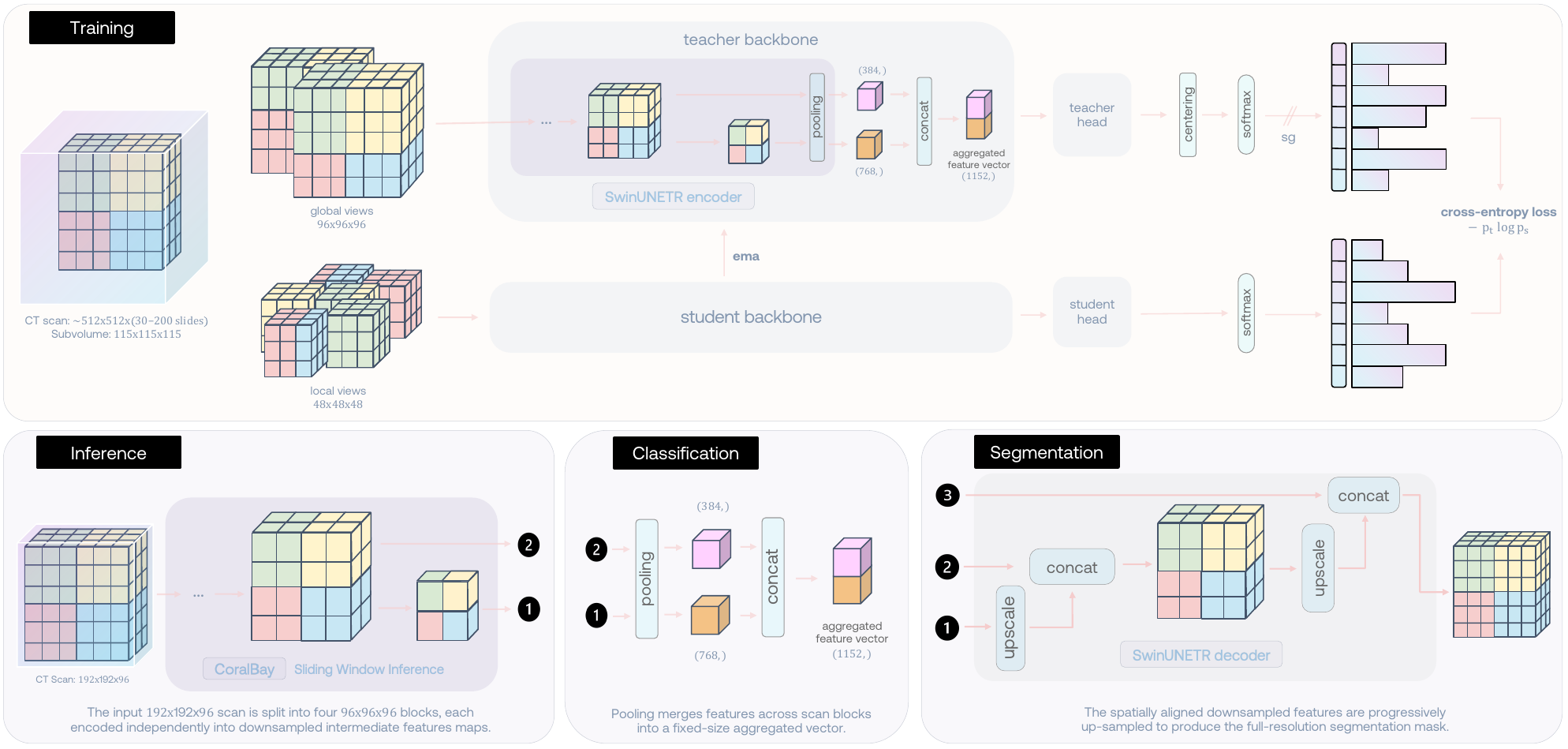}
    \end{subfigure}}
    \captionsetup{font=small}
    \caption[CoralBay]{
        \textbf{Self-supervised training and downstream inference pipeline.}
        \textbf{Top (Training):} A DINO-based distillation framework processes global ($96^3$) and local ($48^3$) crops through student and EMA-updated teacher \textbf{3D SwinTransformer} backbones to minimize distribution cross-entropy loss.
        \textbf{Bottom (Inference):} A \textbf{sliding window technique} processes $96^3$ ROI crops of the scan through the encoder. These downsampled intermediate features are stitched to preserve spatial alignment with the input scan.
        For \textbf{Classification}, a pooling operation merges features into a rich aggregated vector of consistent shape.
        For \textbf{Segmentation}, features are passed to a \textbf{Swin UNETR decoder} with skip connections and upscaling for voxel-wise masks.
    }
    \label{fig:coralbay-pipelines}
\end{figure}

We introduce \coralbay, a self-supervised learning framework that extends the highly effective yet simple DINO self-distillation paradigm from 2D natural images to native 3D volumetric data \cite{caron2021dino}. Our design preserves the simplicity and stability of the original DINO formulation while enabling the learning of rich, hierarchical, and spatially coherent representations from volumetric medical scans. This property is particularly critical for dense prediction tasks in medical imaging, such as organ and lesion segmentation, where both global anatomical context and fine-grained local structure must be captured.

\vspace{-4pt}\paragraph{\textbf{Swin Transformer backbone and multi-resolution features.}}
We adopt a hierarchical Swin Transformer encoder \cite{liu2021swin} following the design principles of Swin UNETR \cite{hatamizadeh2022swinunetr} (figure \ref{fig:coralbay-pipelines}). The shifted-window self-attention mechanism is well suited to volumetric inputs, as it enables efficient modeling of long-range spatial dependencies while maintaining manageable computational and memory costs. Progressive patch merging produces a multi-scale feature hierarchy that naturally captures fine anatomical details—such as vessel boundaries or tumor margins—alongside coarse structural context, including organ shape and spatial relationships. 

For each stage, we apply 3D adaptive average pooling to obtain a fixed-length representation for each resolution, which is then concatenated across resolutions. The resulting feature vector encodes both global anatomical semantics and fine-grained local structure, enabling the standard DINO loss to supervise a scale-aware representation. This design allows us to avoid the compute-intensive iBOT loss component in DINOv2\cite{oquab2024dinov2learningrobustvisual}, while still learning high-resolution local details.

\vspace{-4pt}\paragraph{\textbf{3D volumetric crops and radiology-specific augmentations.}}
To support native 3D processing, \coralbay extends the concept of views from 2D to 3D. Following the local-to-global principle introduced in DINO \cite{caron2021dino}, in each training epoch, from a ct-scan we extract a random volume of size $115\times 115\times115$ (padded if needed), from which we sample two global crops ($96\times96\times96$) and six local crops ($48\times48\times48$). We design an augmentation pipeline tailored to CT imaging, incorporating random contrast adjustments, Gaussian smoothing, and histogram shifts that realistically simulate acquisition variability and reconstruction artifacts \cite{zhou2021modelsgenesis}. A key component of the augmentation pipeline is the random HU windowing: for each crop we randomly sample from multiple clinically relevant HU windows spanning soft tissue, lung, abdomen, liver, brain, and full-range CT views (Table \ref{tab:hu_ranges}). This strategy encourages the encoder to learn representations that are invariant to windowing choices and robust across anatomical regions, enabling adaptation to diverse downstream tasks without fine-tuning.

\begin{table}[t]
\centering
\captionsetup{justification=centering, skip=4pt, font=small}
\caption{HU ranges for pre-training data augmentation.}
\label{tab:hu_ranges}
\scalebox{0.95}{
\begin{tabular}{r@{\hspace{0.3cm}}l}
\toprule
\textbf{HU Range} & \textbf{Applicable for:}\\ 
\midrule
(-1024, 1024) & Full CT Range (air, soft tissue, bone, metal, etc.) \\
(-1000, 1000) & Whole Body (excluding high-density bone/metal) \\
(-1024, 300)  & Soft Tissues + Lungs (air, lungs, fat, muscle, some soft tissues) \\
(-1000, 600)  & Soft Tissues + Partial Bone (air, fat, muscle, partial bone) \\
(-175, 250)   & Chest / Lung Window (lung parenchyma, soft tissues, blood vessels) \\
(-125, 275)   & Abdomen / Soft Tissue (fat, organs, muscle) \\
(-57, 164)    & Liver / Soft Tissue Focus (liver, spleen, kidney, muscle) \\
(-21, 189)    & Brain / Soft Tissue (gray/white matter, cerebrospinal fluid) \\
\bottomrule
\end{tabular}
}
\end{table}

\vspace{-4pt}\paragraph{\textbf{Scan level inference.}} While the Swin transformer backbone is trained to work with crops of size around $96\times96\times96$, we apply the sliding window technique at inference to obtain scan level features. A scan is divided into crops of the shape $96\times96\times96$, which are encoded by the backbone independently. The crop level representations are further stitched (and pooled) together as the scan level representation for downstream tasks.

\vspace{-4pt}\paragraph{\textbf{Data.}}
To develop a robust and generalizable model, we constructed from multiple publicly available sources a large-scale, balanced collection of medical imaging volumes that adequately represents all major anatomical regions, termed \texttt{CORID} (\textbf{C}ombination \textbf{O}f \textbf{R}adiology \textbf{I}mage \textbf{D}ata). The collection proportionally represents all major anatomical regions, such as the chest, abdomen, lung, and head \& neck, was proportionally represented, minimizing potential biases towards any single modality or anatomical area. Table~\ref{tab:pretrained_datasets} summarizes the included datasets.

\begin{table}[b]
\centering
\captionsetup{justification=centering, skip=4pt, font=small}
\caption{Summary of pre-training datasets; Later versions are supersets of earlier versions.}
\label{tab:pretrained_datasets}
\scalebox{0.9}{
\begin{tabular}{l@{\hspace{0.25cm}}c@{\hspace{0.25cm}}c@{\hspace{0.25cm}}c@{\hspace{0.25cm}}l}
\toprule
\textbf{Dataset} & \textbf{Volumes} & \textbf{Region} & \textbf{CORID} & \textbf{Description} \\
\midrule
AbdomenAtlas Mini 1.0 & 5,195 & Abdomen & v1 & Multi-organ Segmentation \\
HNSCC & 627 & Head \& Neck & v2 & Radiotherapy Planning \\
LUNA16 & 888 & Lung & v2 & Lung Nodule Analysis \\
STOIC 2021 & $\sim$2,000 & Chest & v2 & COVID-19 Severity \\
LIDC-IDRI & 1,018 & Lung & v3 & Lung Nodule Detection \\
Stony Brook & 1,384 & Chest & v3 & COVID-19 (Pos. Cases) \\
TCGA-HNSC & 227 & Head \& Neck & v3 & Genomic/Radiomic Analysis \\
\bottomrule
\end{tabular}
}
\end{table}

\vspace{-4pt}\paragraph{\textbf{Training configuration.}}
Using the \coralbay framework, we trained two models of different sizes: \texttt{CoralBayU96B} with $53.2$M parameters and \texttt{CoralBayU96H} with $847$M parameters. Both were trained for 2{,}000 epochs on the \texttt{CORID} dataset using the AdamW optimizer with a cosine learning-rate schedule. The learning rate was scaled linearly with the effective batch size, using a base learning rate of $5 \times 10^{-4}$ for batch size $6$. We used a multi-crop strategy with $6$ local views per sample, global crops of size $96\times96\times96$ frames, and local crops of size $48\times48\times48$ frames. Pre-training included a $50$-epoch learning rate warmup period, during which the last layer of the DINO projection head were frozen for the first $3$ epochs. For the DINO loss, we used a teacher temperature of $0.03$.


\section{Results}

To evaluate the efficacy of the \coralbay framework, we benchmark \texttt{CoralBayU96B} and \texttt{CoralBayU96H} on 11 diverse datasets spanning both global and fine-grained tasks. We assess scan-level classification to measure holistic understanding, alongside multi-organ and small-lesion segmentation across multiple anatomical regions to evaluate precise localization and delineation capabilities. Quantitative results are reported in Table~\ref{tab:results}.\footnote{\scriptsize \url{https://github.com/kaiko-ai/eva}}

\begin{table*}[t!]
\arrayrulecolor[rgb]{0.82,0.82,0.82}
\centering
\captionsetup{skip=8pt, font=small}
\caption{Quantitative performance across classification (Multiclass Accuracy/Binary AUROC) and segmentation (Dice score) tasks, as evaluated via the \eva{} framework.}
\label{tab:results}
\setlength{\tabcolsep}{4pt}
\renewcommand{\arraystretch}{1.35}
\scalebox{0.8}{
\begin{tabular}{lcc @{\hspace{1.5em}} cccc @{\hspace{1.5em}} ccccccc}
\multirow{2}{*}{\textbf{Model}} & &
\multirow{2}{*}{\shortstack{\textbf{\# pre-trained}\\\textbf{data}}} &
\rotatebox{90}{\textbf{NoduleMNIST}} &
\rotatebox{90}{\textbf{OrganMNIST}} &
\rotatebox{90}{\textbf{CC-CCII}} &
\rotatebox{90}{\textbf{LUNA25}}&
\rotatebox{90}{\textbf{BTCV}} &
\rotatebox{90}{\textbf{WORD}} &
\rotatebox{90}{\textbf{FLARE22}} &
\rotatebox{90}{\textbf{CHAOS}} &
\rotatebox{90}{\textbf{LiTs17}} &
\rotatebox{90}{\textbf{KiTs23}} &
\rotatebox{90}{\textbf{MSD Pancreas}} \\
& & &
\multicolumn{4}{c@{\hspace{1.5em}}}{\cellcolor[gray]{.96}\textbf{Classification}} &
\multicolumn{7}{c}{\cellcolor[gray]{.96}\textbf{Segmentation}} \\[0.0ex]

SwinUNETR & \raisebox{-.2ex}{\includegraphics[height=1.2ex]{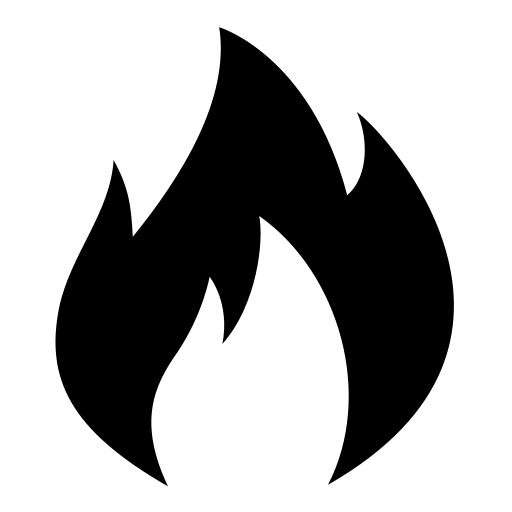}} & ---
  & .83 & .99 & .90 & .81
  & .74 & .85 & .90 & .96
  & .82 & .74 & .72 \\[-1ex]

nnUNetV2 & \raisebox{-.2ex}{\includegraphics[height=1.2ex]{assets/fire.png}} & ---  
  & --- & --- & --- & ---  
  & .85 & --- & --- & --- & .84 & .81 & .75 \\[0.0ex]

Universal Model \cite{Liu_2023} & \raisebox{-.2ex}{\includegraphics[height=1.2ex]{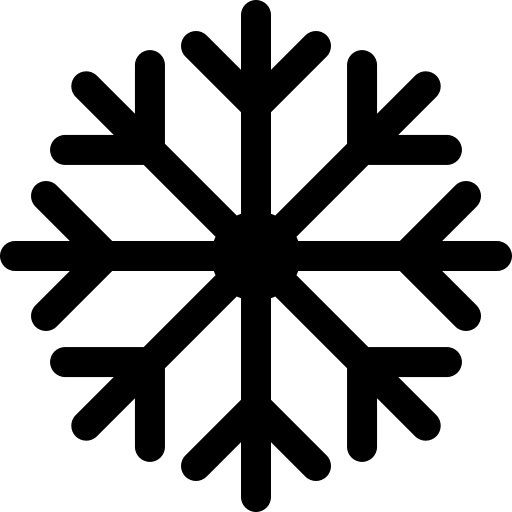}} & 2.1k
  & .78 & .96 & .77 & .72
  & .81 & .83 & .91 & \textbf{.97}
  & .81 & .74 & .67 \\[-1ex]

SuPreM \cite{li2025supervised3dmodelstransfer} & \raisebox{-.2ex}{\includegraphics[height=1.2ex]{assets/frozen.png}} & 2.1k
  & .71 & .98 & .83 & \textbf{.74}
  & .81 & .85 & .91 & \textbf{.97}
  & .81 & .80 & \textbf{.72} \\[-1ex]

VoCo-B \cite{voco} & \raisebox{-.2ex}{\includegraphics[height=1.2ex]{assets/frozen.png}} & 160k  
  & .74 & .96 & .71 & .72  
  & \textbf{.83} & \textbf{.86} & \textbf{.92}
  & \textbf{.97} & .81 & .78 & .71 \\[-1ex]

VoCo-H \cite{voco} & \raisebox{-.2ex}{\includegraphics[height=1.2ex]{assets/frozen.png}} & 160k  
  & .76 & .92 & .83 & .70  
  & \textbf{.83} & .85  & \textbf{.92} & \textbf{.97}
  & \textbf{.82} & \textbf{.82} & \textbf{.72} \\[0.5ex]

\rowcolor[gray]{0.96}
CoralBayU96B & \raisebox{-.2ex}{\includegraphics[height=1.2ex]{assets/frozen.png}} & 11k  
  & .79 & \textbf{.99} & .76 & .72
  & .81 & .85 & .91 & \textbf{.97}
  & .78 & .79 & .65 \\[-0.5ex]

\rowcolor[gray]{0.96}
CoralBayU96H & \raisebox{-.2ex}{\includegraphics[height=1.2ex]{assets/frozen.png}} & 11k  
  & \textbf{.80} & \textbf{.99} & \textbf{.90} & \textbf{.74}
  & .82  & .85 & .91 & \textbf{.97}
  & .81 & .81 & .67 \\[-0.5ex]

\rowcolor[gray]{0.96}
CoralBayU96H \color[HTML]{9B9B9B} \texttt{(FT)} & \raisebox{-.2ex}{\colorbox[gray]{0.96}{\includegraphics[height=1.2ex]{assets/fire.png}}} & 11k  
  & .91 & 1.0 & .91 & .82
  & .83 & .86 & .91 & .97
  & .84 & .82 & .73 \\

\end{tabular}
}
\end{table*}

\subsection{Classification}

We evaluate scan-level representation quality through linear probing across four 3D CT benchmarks: multiclass organ identification (OrganMNIST3D), lung nodule malignancy prediction (NoduleMNIST3D, LUNA25), and COVID-19 pathology classification (CC-CCII). We report Multiclass Accuracy and Binary AUROC to account for class imbalances.

Across the four classification benchmarks, \coralbay demonstrates consistently strong performance, with CoralBayU96H achieving the best or tied-best results. Notably, it matches or exceeds large-scale SSL models despite substantially fewer pretraining samples, indicating improved linear separability and enhanced sensitivity to fine-grained pathological features. These results suggest that \coralbay's representations preserve discriminative cues critical for pathology-centric tasks, rather than encoding task-irrelevant anatomical bias, leading to robust generalization across diverse classification settings. The model's high-resolution feature extraction is further validated by its strong performance on the LUNA25 malignancy prediction task and its competitive $0.93$ AUROC on the official open development leaderboard\footnote{\scriptsize \url{https://luna25.grand-challenge.org/evaluation/open-development-phase/leaderboard/}}, achieved without the need for task-specific fine-tuning or complex ensembling strategies.

\subsection{Segmentation}

Unless stated otherwise, we evaluate segmentation performance using the average sample-wise macro-Dice score. To evaluate representation quality, we adopt a segmentation analogue of linear probing by freezing the pretrained encoder and training only the Swin UNETR decoder~\cite{hatamizadeh2022swinunetr}, adding 3D $1 \times 1 \times 1$ convolutions to downsample encoder features and reduce the decoder size from 313M to 22.8M parameters, ensuring a lightweight and consistent capacity across variants.


As shown in Table~\ref{tab:results}, \coralbay exhibits strong data efficiency, performing comparably to VoCo despite using $<7\%$ of its label-free pretraining data. Unlike Universal Model and SuPreM which use labeled pretraining, \coralbay excels at fine-grained features on challenging tumor datasets (LiTS17: $0.81$, KiTS23: $0.81$). With a frozen encoder it already rivals heavily tuned nnU-NetV2, which typically relies on exhaustive, task-specific heuristics, highlighting robust capture of low-contrast and small lesions. Finaly the full fine-tuning of CoralBayU96H consistently outperforms SwinUNETR and closely matches the nnU-NetV2, validating the proposed pretraining for complex anatomy and pathology.

\subsection{Ablation studies}

\begin{figure}[t]
    \centering
    \begin{minipage}{0.4\textwidth}
        \centering
        \includegraphics[height=5.8cm,width=\linewidth,keepaspectratio]{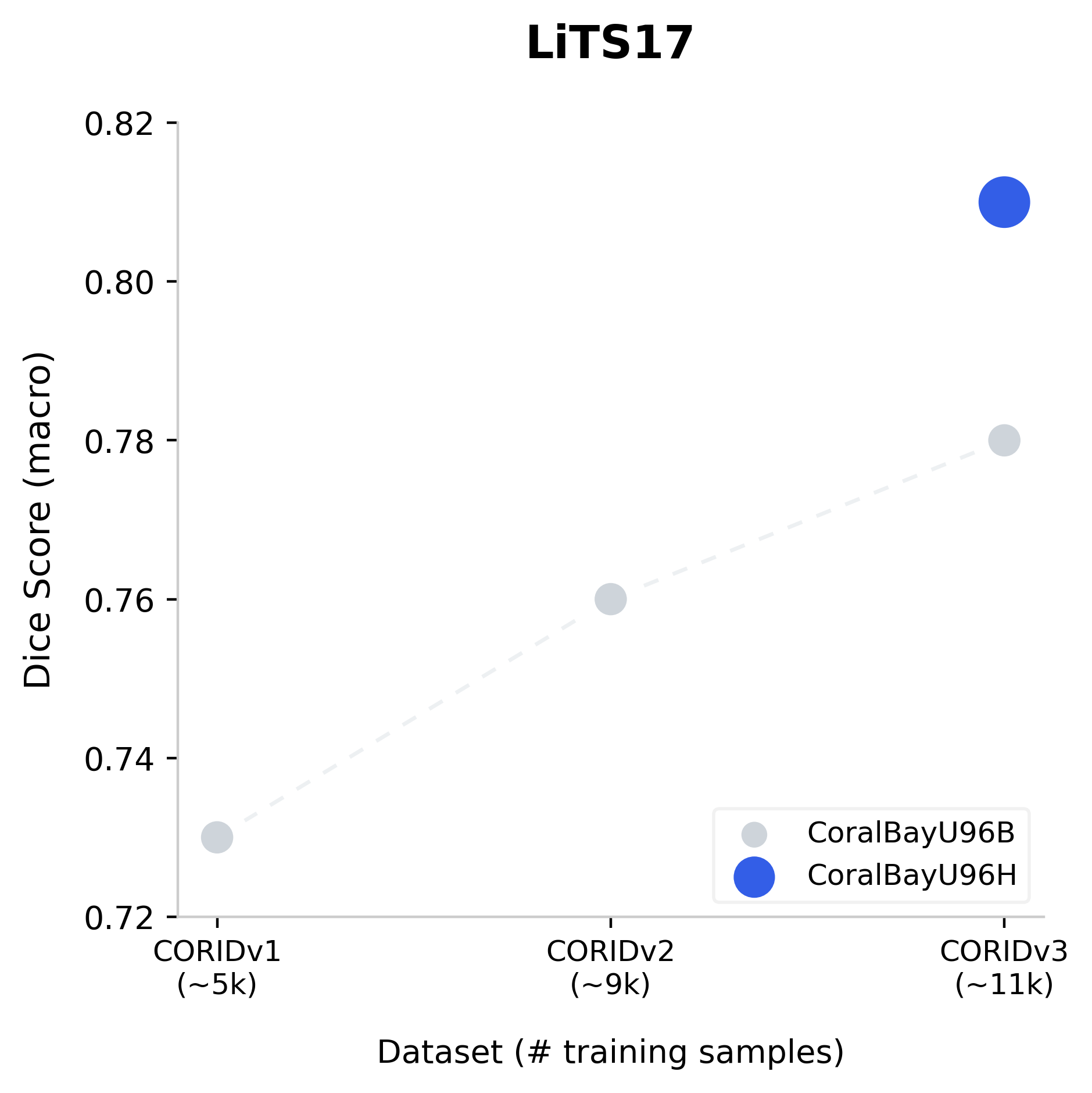}
    \end{minipage}%
    \hfill
    \begin{minipage}{0.6\textwidth}
        \centering
        \includegraphics[height=5.8cm,width=\linewidth,keepaspectratio]{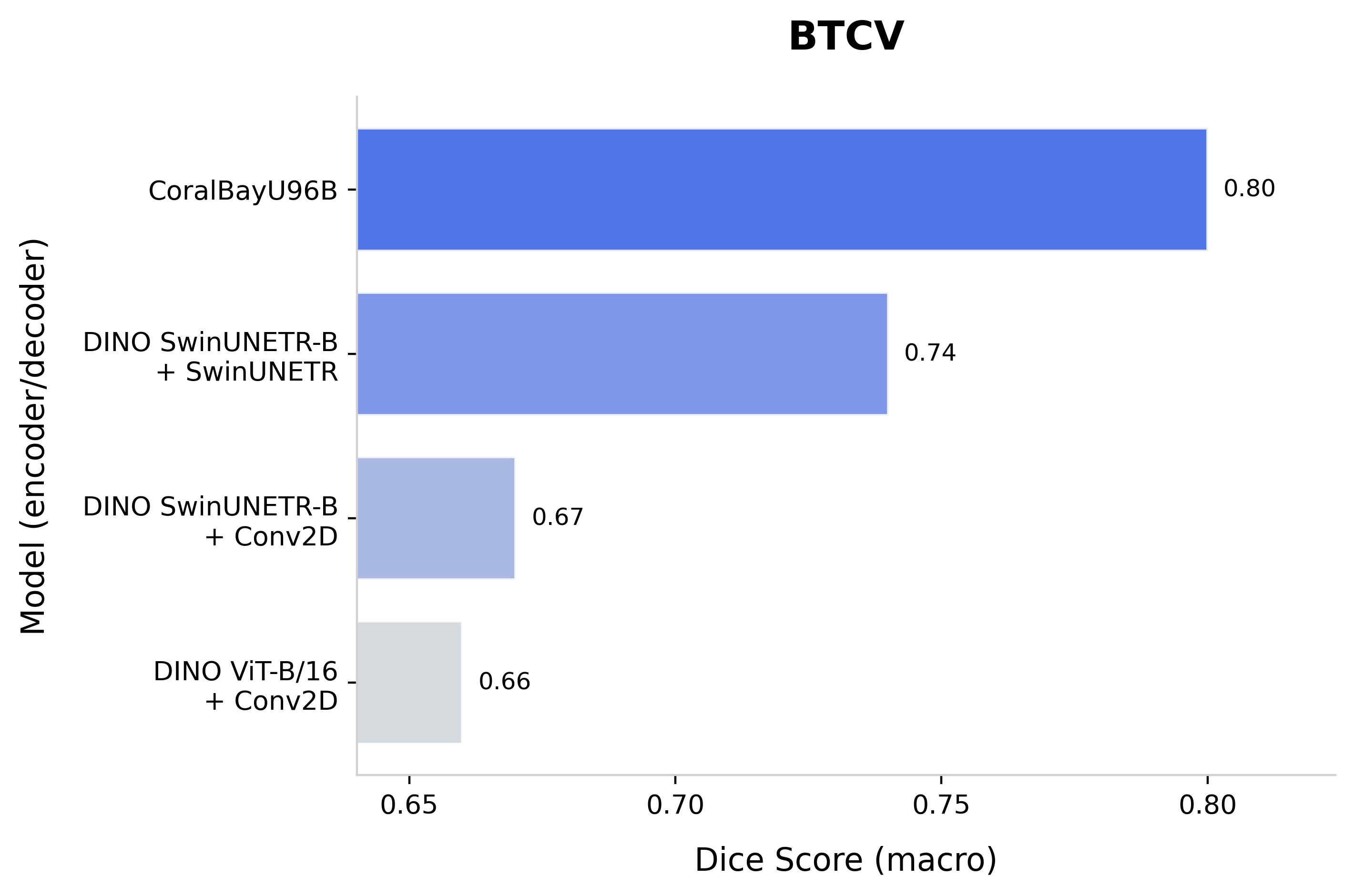}
    \end{minipage}
    \vspace{0.8em}
    \begin{minipage}{0.32\textwidth}
        \centering
        \includegraphics[height=5.8cm,width=\linewidth,keepaspectratio]{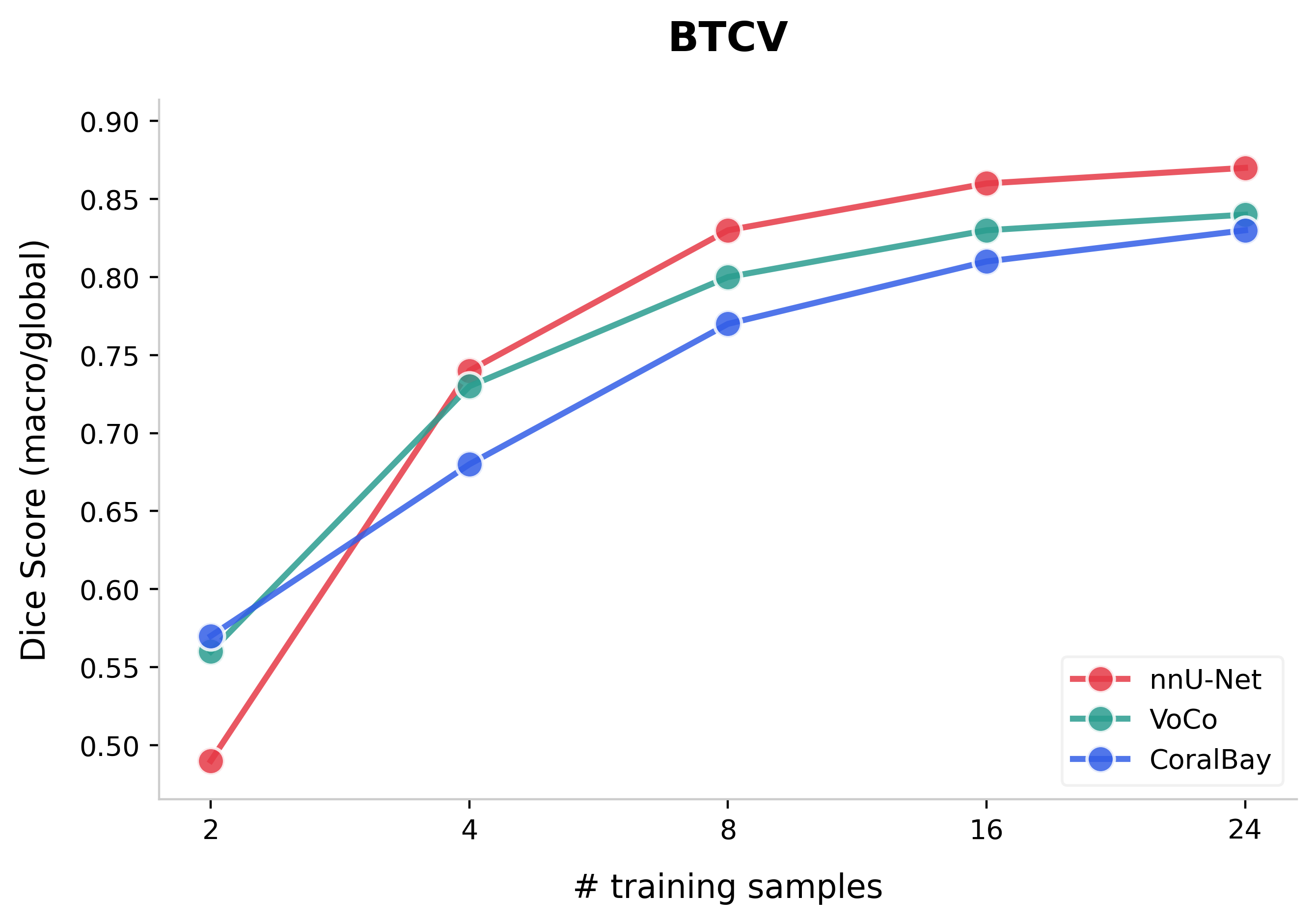}
    \end{minipage}%
    \hfill
    \begin{minipage}{0.32\textwidth}
        \centering
        \includegraphics[height=5.8cm,width=\linewidth,keepaspectratio]{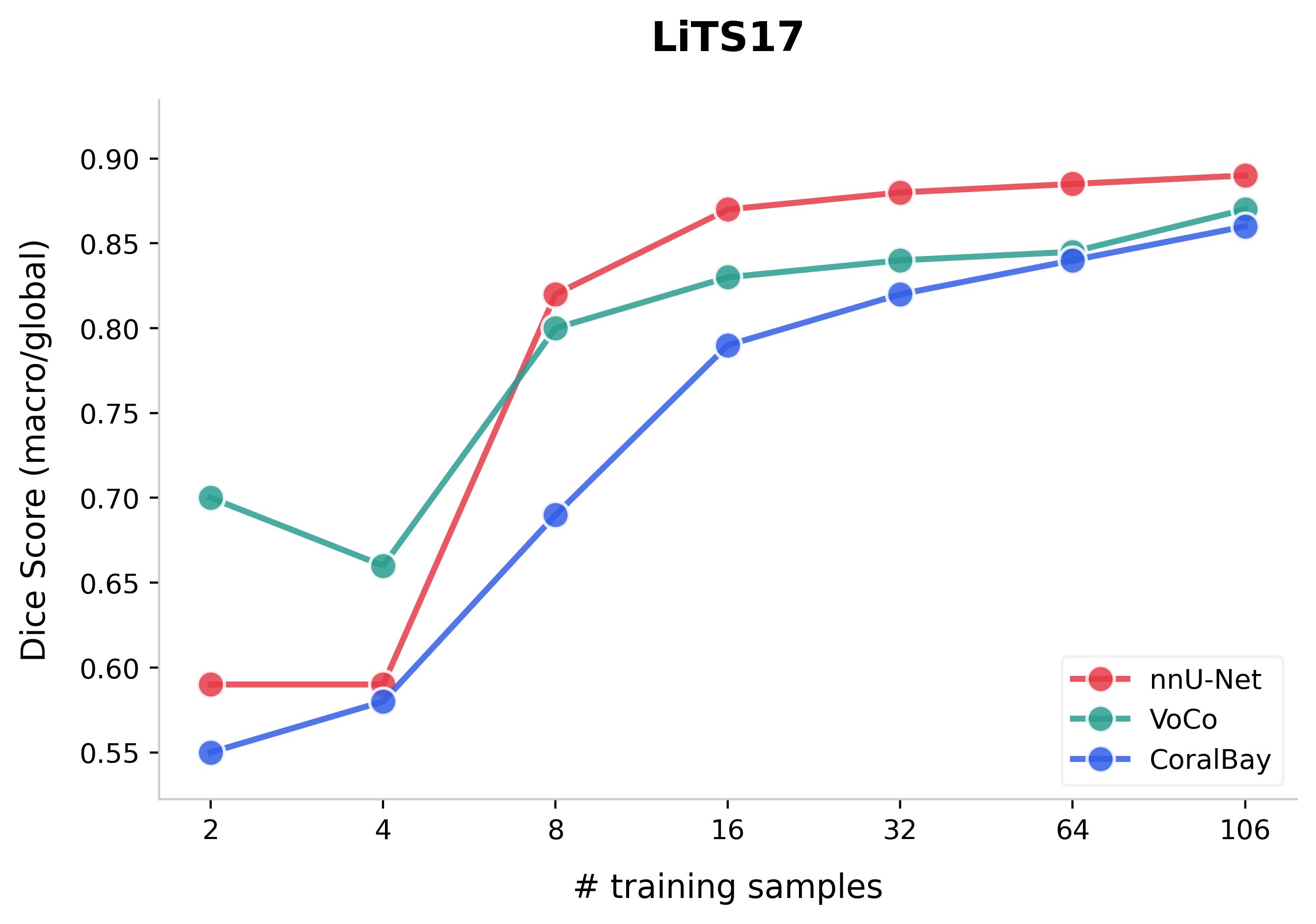}
    \end{minipage}%
    \hfill
    \begin{minipage}{0.32\textwidth}
        \centering
        \includegraphics[height=5.8cm,width=\linewidth,keepaspectratio]{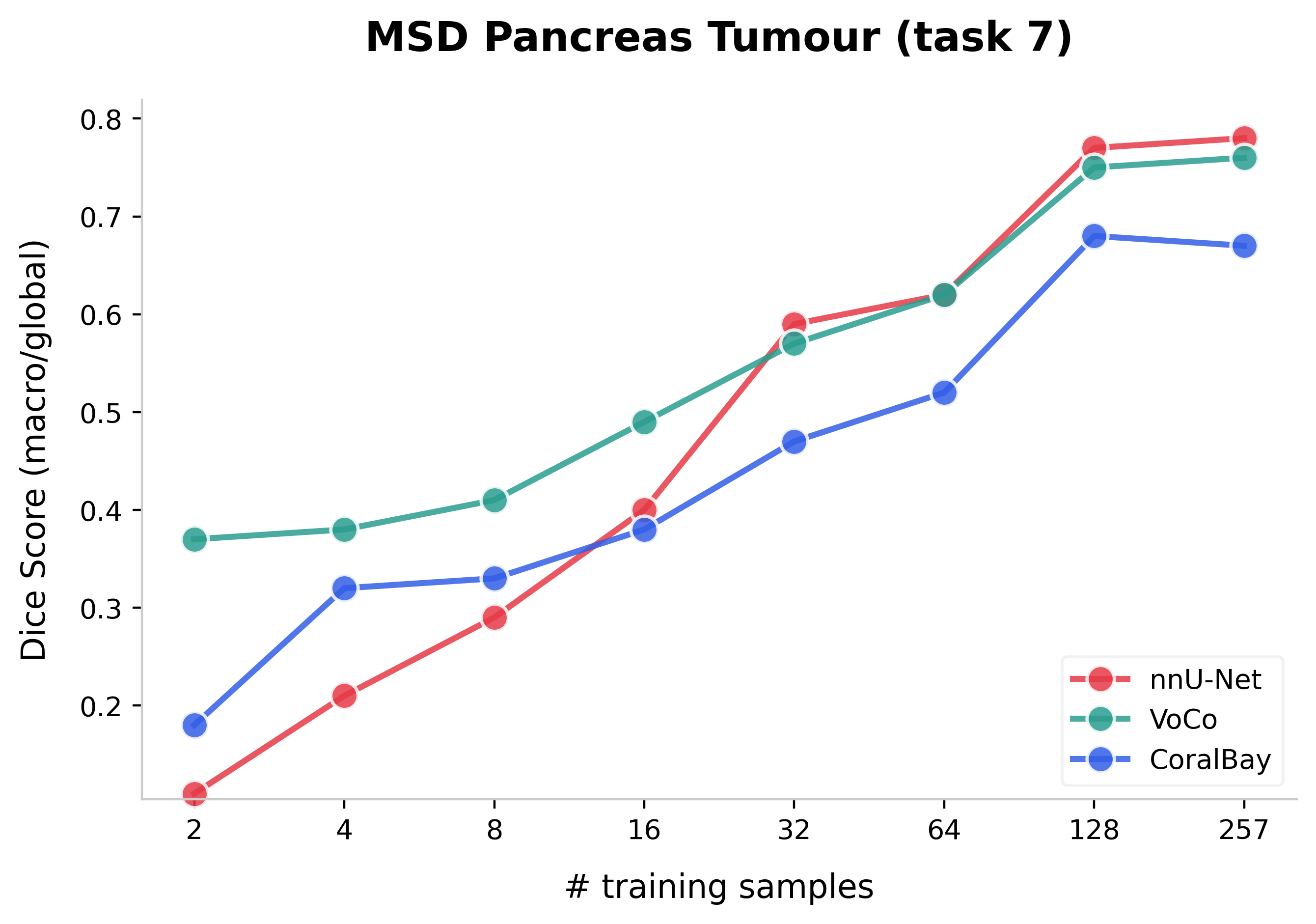}
    \end{minipage}
    \captionsetup{skip=6pt, font=small}
    \caption{Ablation studies. 
    \textbf{Top Left:} Scaling model \& pre-training dataset size.
    \textbf{Top right:} 2D vs.\ 3D encoder-decoder configurations (frozen encoder).
    \textbf{Bottom:} Label efficiency across tasks of increasing difficulty (from left $\rightarrow$ right).}
    \label{fig:ablation_studies}
\end{figure}

\subsubsection{Scaling Model \& Data}
We evaluate \coralbay on \texttt{LiTS17} after pre-training on progressively larger \texttt{CORID} subsets. Segmentation performance improves consistently with dataset size (Figure \ref{fig:ablation_studies}a), showing that the framework scales effectively to larger data and model capacities. These results indicate that increasing both dataset size and model scale enhances downstream segmentation and that pre-training on a balanced dataset promotes strong generalization.

\subsubsection{Label Efficiency}
We assess label efficiency by training all models on progressively smaller subsets across three segmentation tasks of increasing difficulty: BTCV (organ), LiTS17 (liver tumor), and MSD Pancreas Tumor (Figure \ref{fig:ablation_studies}c–e). On the simpler organ task, foundation models show no clear advantage over nnU-Net, which remains highly label-efficient due to extensive augmentations. For the more challenging tumor tasks, both CoralBayU96B and VoCo-B outperform nnU-Net in low-data settings, demonstrating that self-supervised pre-training provides a stronger prior for small, low-contrast pathological structures.


\subsubsection{2D vs 3D Modelling}
Replacing the standard DINO ViT-B/16 2D encoder with a Swin UNETR-B 2D backbone pretrained using DINO within the CoralBay framework yields only a marginal gain when paired with a simple Conv2D decoder ($0.66$ → $0.67$). However, switching to the matching Swin UNETR 3D decoder brings a substantial jump to $0.74$ Dice, demonstrating that most of the benefit arises from 3D spatial reasoning rather than the choice of 2D pre-training alone. The fully 3D-native \texttt{CoralBayU96B} model, pretrained exclusively on the abdominal atlas dataset, further improves performance to $0.80$, highlighting the advantage of consistent 3D inductive biases and volumetric pre-training for abdominal multi-organ segmentation.

\section{Conclusion}

We introduced \coralbay, a self-supervised framework for 3D medical volumes that extends DINO to hierarchical 3D Swin Transformers with multi-resolution features and radiology-specific augmentations. \texttt{CoralBayU96B} ($53.2$M params) and \texttt{CoralBayU96H} ($847$M params) trained with \coralbay with significantly less data achieved strong performance across diverse classification and segmentation tasks. Ablations confirm the benefits of native 3D modeling and multi-scale feature learning. By integrating with the \eva framework and a public 3D segmentation leaderboard, \coralbay provides a reproducible benchmark for future research in volumetric medical imaging.

\bibliographystyle{splncs04}
\bibliography{bibliography}

@inproceedings{luna25,
  title     = {Benchmarking of Artificial Intelligence and Radiologists for Lung Cancer Screening in CT: The LUNA25 Challenge},
  author    = {Peeters, D. and Obreja, B. and Antonissen, N. and Jacobs, C.},
  booktitle = {Proceedings of Medical Image Computing and Computer Assisted Intervention (MICCAI)},
  year      = {2025},
  publisher = {Zenodo},
  doi       = {10.5281/zenodo.15094631},
  url       = {https://doi.org/10.5281/zenodo.15094631}
}

@article{voco,
  title={Large-scale 3d medical image pre-training with geometric context priors},
  author={Wu, Linshan and Zhuang, Jiaxin and Chen, Hao},
  journal={IEEE Transactions on Pattern Analysis and Machine Intelligence},
  year={2025},
  publisher={IEEE}
}

@misc{li2025supervised3dmodelstransfer,
  title={How Well Do Supervised 3D Models Transfer to Medical Imaging Tasks?}, 
  author={Wenxuan Li and Alan Yuille and Zongwei Zhou},
  year={2025},
  eprint={2501.11253},
  archivePrefix={arXiv},
  primaryClass={eess.IV},
  url={https://arxiv.org/abs/2501.11253}
}

@inproceedings{Liu_2023,
  title={CLIP-Driven Universal Model for Organ Segmentation and Tumor Detection},
  url={http://dx.doi.org/10.1109/ICCV51070.2023.01934},
  DOI={10.1109/iccv51070.2023.01934},
  booktitle={2023 IEEE/CVF International Conference on Computer Vision (ICCV)},
  publisher={IEEE},
  author={Liu, Jie and Zhang, Yixiao and Chen, Jie-Neng and Xiao, Junfei and Lu, Yongyi and Landman, Bennett A. and Yuan, Yixuan and Yuille, Alan and Tang, Yucheng and Zhou, Zongwei},
  year={2023},
  month=oct, 
  pages={21095--21107} 
}

@article{bommasani2021opportunities,
  title={On the opportunities and risks of foundation models},
  author={Bommasani, Rishi and Hudson, Drew A and Adeli, Ehsan and others},
  journal={arXiv preprint arXiv:2108.07258},
  year={2021}
}

@inproceedings{he2020momentum,
  title={Momentum contrast for unsupervised visual representation learning},
  author={He, Kaiming and Fan, Haoqi and Wu, Yuxin and Xie, Saining and Girshick, Ross},
  booktitle={Proceedings of the IEEE/CVF Conference on Computer Vision and Pattern Recognition},
  pages={9729--9738},
  year={2020}
}

@article{caron2021dino,
  title={Emerging properties in self-supervised vision transformers},
  author={Caron, Mathilde and Touvron, Hugo and Misra, Ishan and others},
  journal={Proceedings of the IEEE/CVF International Conference on Computer Vision},
  year={2021}
}

@inproceedings{radford2021learning,
  title={Learning transferable visual models from natural language supervision},
  author={Radford, Alec and Kim, Jong Wook and Hallacy, Chris and others},
  booktitle={International Conference on Machine Learning (ICML)},
  pages={8748--8763},
  year={2021}
}

@inproceedings{chen2020simple,
  title={A simple framework for contrastive learning of visual representations},
  author={Chen, Ting and Kornblith, Simon and Norouzi, Mohammad and Hinton, Geoffrey},
  booktitle={International Conference on Machine Learning (ICML)},
  year={2020}
}

@article{jing2020self,
  title={Self-supervised visual feature learning with deep neural networks: A survey},
  author={Jing, Longlong and Tian, Yingli},
  journal={IEEE Transactions on Pattern Analysis and Machine Intelligence},
  volume={43},
  number={11},
  pages={4037--4058},
  year={2020}
}

@article{taleb20213d,
  title={3D self-supervised methods for medical imaging},
  author={Taleb, Aiham and Lippert, Christoph and Klein, Tassilo and Nabi, Moin},
  journal={Advances in Neural Information Processing Systems (NeurIPS)},
  year={2021}
}

@article{zhang2022self,
  title={Self-supervised pretraining of 3D medical image models by learning region-aware representations},
  author={Zhang, Yuyin and others},
  journal={Medical Image Analysis},
  year={2022}
}

@article{grill2020bootstrap,
  title={Bootstrap your own latent: A new approach to self-supervised learning},
  author={Grill, Jean-Bastien and Strub, Florian and Altch{\'e}, Florent and others},
  journal={Advances in Neural Information Processing Systems (NeurIPS)},
  year={2020}
}

@article{zhou2021modelsgenesis,
  title={Models Genesis: Generic autodidactic models for 3D medical image analysis},
  author={Zhou, Zongwei and Sodha, Vatsal and Pang, Jiaxuan and Gotway, Michael B. and Liang, Jianming},
  journal={Medical Image Analysis},
  volume={67},
  pages={101840},
  year={2021}
}

@article{hatamizadeh2022swinunetr,
  title={Swin UNETR: Swin transformers for semantic segmentation of brain tumors in MRI images},
  author={Hatamizadeh, Ali and Tang, Yucheng and Nath, Vignesh and others},
  journal={Proceedings of the IEEE/CVF Conference on Computer Vision and Pattern Recognition (CVPR)},
  year={2022}
}

@article{shen2017deep,
  title={Deep learning in medical image analysis},
  author={Shen, Dinggang and Wu, Guorong and Suk, Heung-Il},
  journal={Annual Review of Biomedical Engineering},
  volume={19},
  pages={221--248},
  year={2017}
}

@misc{oquab2024dinov2learningrobustvisual,
  title={DINOv2: Learning Robust Visual Features without Supervision}, 
  author={Maxime Oquab and others},
  year={2024},
  eprint={2304.07193},
  archivePrefix={arXiv},
  primaryClass={cs.CV}
}

@inproceedings{liu2021swin,
  title={Swin transformer: Hierarchical vision transformer using shifted windows},
  author={Liu, Ze and Lin, Yutong and Cao, Yue and Hu, Han and Wei, Yixuan and Zhang, Zheng and Lin, Stephen and Guo, Baining},
  booktitle={Proceedings of the IEEE/CVF International Conference on Computer Vision (ICCV)},
  pages={10012--10022},
  year={2021}
}

@inproceedings{tang2022self,
  title={Self-supervised pre-training of swin transformers for 3d medical image analysis},
  author={Tang, Yucheng and Yang, Dong and Li, Wenqi and Roth, Holger R and Landman, Bennett A and Xu, Daguang and Nath, Vishwesh and Hatamizadeh, Ali},
  booktitle={Proceedings of the IEEE/CVF Conference on Computer Vision and Pattern Recognition (CVPR)},
  pages={20730--20740},
  year={2022}
}

@inproceedings{gatopoulos2024eva,
  title={eva: Evaluation framework for pathology foundation models},
  author={Gatopoulos, Ioannis and K{\"a}nzig, Nicolas and Moser, Roman and Ot{\'a}lora, Sebastian and others},
  booktitle={Medical Imaging with Deep Learning},
  year={2024}
}

\end{document}